\documentclass[runningheads]{llncs}
\usepackage{graphicx}
\usepackage{comment}
\usepackage{multirow}
\usepackage[hyphens]{url}
\usepackage{booktabs}
\usepackage{xurl}

\usepackage{subcaption}

\begin{document}
\title{A Review of Benchmarks for Visual Defect Detection in the Manufacturing Industry}
\titlerunning{Visual Defect Detection in the Manufacturing Industry}
%
\author{Philippe Carvalho\inst{1} \and
Alexandre Durupt\inst{1} \and
Yves Grandvalet\inst{2}}
\authorrunning{P. Carvalho et al.}
%
\institute{Roberval, Université de technologie de Compiègne, France\\
           \email{\{philippe.carvalho,alexandre.durupt\}@utc.fr}\and
           Heudiasyc, UMR CNRS 7253, Université de technologie de Compiègne, France\\
           \email{yves.grandvalet@utc.fr}}
\maketitle              
\begin{abstract}

The field of industrial defect detection using machine learning and deep learning is a subject of active research. Datasets, also called benchmarks, are used to compare and assess research results. There is a number of datasets in industrial visual inspection, of varying quality. Thus, it is a difficult task to determine which dataset to use. Generally speaking, datasets which include a testing set, with precise labeling and made in real-world conditions should be preferred. We propose a study of existing benchmarks to compare and expose their characteristics and their use-cases. A study of industrial metrics requirements, as well as testing procedures, will be presented and applied to the studied benchmarks. We discuss our findings by examining the current state of benchmarks for industrial visual inspection, and by exposing guidelines on the usage of benchmarks.


\end{abstract}
%
%
%




\section{Introduction}
\label{section:introduction}

\begin{figure}[h]
    \centering
    \includegraphics[scale=0.4]{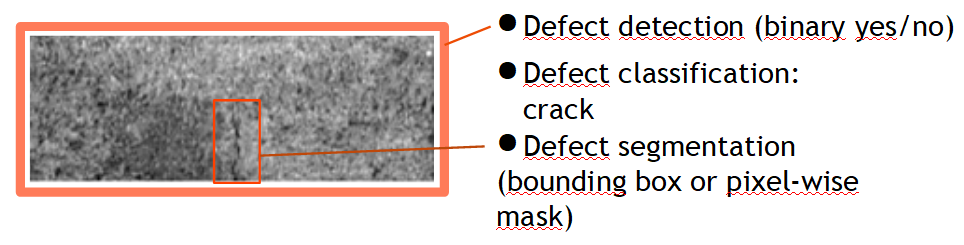}
    \caption{Defect detection, classification and segmentation examples. Sample taken from the KolektorSDD dataset \cite{Tabernik2020}}
    \label{fig:defect_detect}
\end{figure}

The field of industrial defect detection is a subject of intense research. Most publications present new datasets and new algorithms evaluated using industrial defect datasets.
Gao et al. \cite{Gao2021} and Chen et al. \cite{Chen2021} present two reviews on recent advances in industrial defect detection, which show that a large number of algorithms have been tested on a large number of datasets. This process of benchmarking, which consists of comparing a certain number of algorithms on a common dataset, using the same metrics, is important because it gives a fair estimate of the performance of an algorithm, and allows comparison with other algorithms. Several kinds of benchmarks can be made: defect detection (detecting whether an image contains a defect), defect classification (identify which type of defect it is) and defect segmentation (identify the boundaries of the defect on the picture) (cf. Figure \ref{fig:defect_detect}). In our work, we focus on defect detection in the production of metallic and plastic parts.

In Section \ref{section:performance_assessment}, we will study how performance is assessed for benchmarking and what results they depict. In Section \ref{section:review}, we will review and compare the datasets most usually used in visual defect detection benchmarks. Section \ref{section:discussion} will discuss the main findings of this article and present our guidelines regarding the current state of the benchmarks in industrial defect detection.

\section{Industrial Needs for Visual Defect Detection}
\label{section:performance_assessment}

The detection of defects is essential in the manufacturing industry, since the presence or absence of a defect determines the validity of the analyzed item. We will refer to this task as \textit{defect detection}.
It may also be useful to determine the type of defect detected, which we will refer to as \textit{defect classification}. This can be useful for gathering statistics on the types of defects most often encountered, and for performing a diagnosis on the manufacturing process.
Finally, it is also useful to retrieve the location of the defect in the image, which we will refer to as \textit{defect segmentation}. This location makes it easy to check the result of the defect detection.

\subsection{Metrics for classification}
The metrics used in visual defect detection and classification often originate from the field of machine learning, namely precision, recall and F-score. Other measures can be derived from a confusion matrix but are seldom used. In industrial visual inspection, defect detection algorithms must be able to detect any kind of defect, including defects that have not been seen by the classifier during training, and do so on a real-time production line. Other measures can be introduced here to better represent the performance of these algorithms in this setting : the false positive rate (FPR) and the false negative rate (FNR). The accuracy measure, which is sometimes used to describe the performance of a machine learning algorithm, is not relevant in industrial defect detection due to the strong imbalance between classes.

\begin{table}[]
    \centering
    \caption{Confusion matrix used to gather evaluation statistics}
    \label{tab:evaluation_statistics}
    \begin{tabular}{|c|c|c|c|}
         \cline{3-4}
         \multicolumn{2}{c|}{}  & \multicolumn{2}{c|}{Ground Truth} \\ \cline{3-4}
         \multicolumn{2}{c|}{}  & Positive & Negative \\ \hline
        \multirow{2}{*}{Prediction} & Positive & True Positives & False Positive \\  \cline{2-4}
         & Negative & False Negative & True Negative \\ \hline
    \end{tabular}
\end{table}

The aforementioned statistics are calculated as follows: \\

\noindent
\begin{equation}
    \textrm{False Positive Rate (FPR)} = \frac{FP}{FP+TN}
\end{equation}
\begin{equation}
    \textrm{False Negative Rate (FNR)} = \frac{FN}{TP+FN}
\end{equation}
\begin{equation}
    \textrm{Precision} = \displaystyle\frac{TP}{TP+FP}
\end{equation}
\begin{equation}
    \textrm{Recall} = \frac{TP}{TP+FN} = 1-FNR
\end{equation}
\begin{equation}
    \textrm{F-score} = 2 \frac{Pr\, Re}{Pr+Re} = \frac{2\, TP}{2\, TP+FP+FN}
\end{equation}
\begin{equation}
    \textrm{Accuracy} = \frac{TP+TN}{TP+TN+FP+FN}
\end{equation}

In this context, positives refer to the rare events, that is, defective items, and negatives refer to the majority class, that is, non-defective items. This is the vocabulary used in most articles as well as in the field of statistics.

Measures such as the average precision (AP) or the Area Under the Receiver-Operator Curve (AUROC, or AUC), although very popular, do not give a proper estimate of the performance of a decision rule. Rather, these measures estimate the performance of a decision rule when modifying its decision threshold, which makes it a measure of a family of decision rules. Furthermore, the AUROC is an inadequate  estimator in samples that contain very few positive examples. Nevertheless, these metrics can still be used to choose the best classifier threshold.

\subsection{Metrics for segmentation}

Segmentation performance is not as simple to evaluate as classification performance: a classification is either correct or incorrect, while a segmentation that is not pixel perfect can still be considered satisfactory.

Many measures can be defined to evaluate a segmentation. Like in the classification case, the accuracy, defined as the ratio of correctly classified positive and negative pixels over all pixels, can be used. The estimator is heavily biased because of class imbalance and should never be used to measure segmentation performance.

Zavrtanik et al. \cite{Zavrtanik2021a} use the average pixel-wise AUROC and AP to estimate segmentation performance on the MVTec dataset \cite{Bergmann2019}. Tabernik et al. \cite{Tabernik2020} again use pixel-wise average precision. Lei et al. \cite{Lei2021} use Intersection over Union (IoU) to determine the segmentation performance.

The metrics based on AUROC and AP are, again, measures of a family of predictors, that can help choose a specific threshold for the predictor. Many other metrics may be used for segmentation, such as the Dice index, popular in the medical image segmentation field \cite{Eelbode2020OptimizationIndex}.

The metrics are computed as such, using $G$ as the label ground truth and $P$ as the predicted pixels:

\begin{equation}
    \textrm{Dice}=\frac{2|G\cap P|}{|G|+|P|}
\end{equation}
\begin{equation}
    \textrm{IoU}=\frac{|G\cap P|}{|G\cup P|}
\end{equation}
\begin{equation}
    \textrm{Boundary IoU}=\frac{|(G_d\cap G)\cap (P_d \cap P)|}{|(G_d \cap G) \cup (P_d \cap P)|}
\end{equation}
where $G_d$ is the pixels located within a distance $d$ from the border of the ground truth segmentation, and $P_d$ is the pixels located within a distance $d$ from the border of the prediction segmentation.

\subsection{Supervision level}
Another fact to take into account is the supervision level during training. Some algorithms, both in segmentation and classification, need stronger supervision signals for training, that is, need more comprehensive labels. In the manufacturing defect detection community, as discussed in Bozic et al. \cite{Bozic2021}, the supervision levels are defined as follows:

\begin{itemize}
    \item Strong supervision: refers to training protocols using positive and negative samples, including the locations of the defects in positive samples. The location is usually given on a separate image file as a mask over the defect in the original image. This mask can be very precise (pixel-wise overlay over the defect) or more weakly defined (bounding box around the defect, such as the ellipses given in the DAGM dataset \cite{Wieler2007}).
    \item Mixed supervision: refers to training protocols using positive and negative samples. Some of the positive samples give the precise defect location, others do not. They are only trained using a boolean index, indicating whether the image is defective or not. 
    \item Weak supervision: refers to training protocols using positive and negative samples. No defect location is given here: all labels are boolean (defective / non-defective).
    \item No supervision: refers to training protocols using only negative samples. The objective is to be able to detect defects during testing, despite none have ever be seen in training.
\end{itemize}

\subsection{Training and evaluation procedure}

Finally, an important part of performance assessment lies in the procedure used for training the model and evaluating it. If the procedures are not harmonized between different articles, results cannot be compared and the benchmarks comparing several algorithms on a given dataset will be flawed. A valid evaluation procedure, in an industrial context, requires data that has never been seen during training to estimate its performance on data that it will see in the future.

A training procedure usually consists of splitting a dataset in three parts, referred to as the training set (usually between 50\% and 70\% of the data), the validation set (15\% to 30\% of the data) and the testing set (15\% to 30\% of the data). These percentages may vary beyond these bounds -- the DAGM testing set contains 50\% of the data.

\begin{itemize}
    \item The training set is used to directly train the model.
    \item The validation set is used for a first estimation of the performance of the model. It is mostly used for comparing different options and to select one.
    \item The test set is used only once, when all the options of the algorithm have been selected. It is used to evaluate the model globally (including option selections) on a never-before-seen data set.
\end{itemize}

Such a procedure ensures that the published results provide an unbiased estimate of the performance of the model on test data. This test data is analogous to the data to those on which the model will have to detect defects on production lines. The defects that will then be observed will all be previously unseen to the classifier: that is why it is particularly important to test performance on unseen data.

Published datasets should provide distinct training and testing sets, to ensure the repeatability and comparability of the results obtained when testing an algorithm.

\section{Review of Existing Benchmarks}
\label{section:review}


A number of datasets have been created in order to train and test algorithms in manufacturing defect detection and estimate their performance. It is important to have a great variety of datasets, because there are a lot of different industries where defects do not have the same properties at all.

In order to compile the following list of datasets, we have decided to focus on industrial images showing metallic or plastic parts in a production line.

We have started by analyzing the DAGM dataset, used for the thesis work of Dekhtiar, 2019 \cite{Dekhtiar2019}, and already presented as popular for industrial defect detection. We have found several dozens of articles using this dataset, and analyzed the other datasets that these articles were using, if they were compatible with our work. We have decided not to keep datasets which were referenced once or twice, which are often private datasets.

We have compared our findings with the work of Chen et al \cite{Chen2020}. While the list of datasets given is much larger, most of the datasets given are not compatible with our focus, and did not achieve sufficient popularity to impact the publications of the field and will thus not be analyzed here.

We will present a succinct description of each dataset. Table \ref{table:datasets} presents and compares the characteristics and use-cases of each dataset.

\begin{figure}[h]
\begin{subfigure}{.5\textwidth}
  \centering
  \includegraphics[width=.3\linewidth]{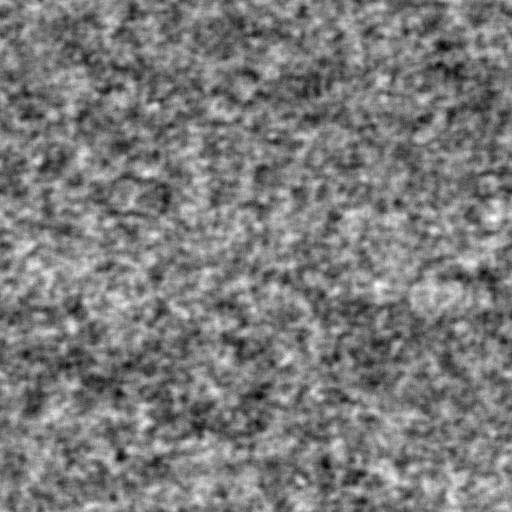}
  \caption{DAGM dataset -- class 9}
  \label{fig:sfig1}
\end{subfigure}%
\begin{subfigure}{.5\textwidth}
  \centering
  \includegraphics[width=.6\linewidth]{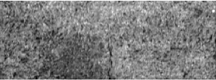}
  \caption{KolektorSDD}
  \label{fig:sfig2}
\end{subfigure}
\begin{subfigure}{.5\textwidth}
  \centering
  \includegraphics[width=.6\linewidth]{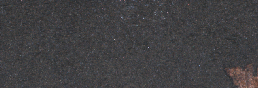}
  \caption{KolektorSDD2}
  \label{fig:sfig2}
\end{subfigure}
\begin{subfigure}{.5\textwidth}
  \centering
  \includegraphics[width=.3\linewidth]{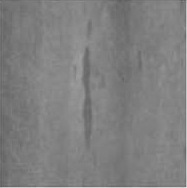}
  \caption{NEU Surface Defect Dataset}
  \label{fig:sfig2}
\end{subfigure}
\begin{subfigure}{.5\textwidth}
  \centering
  \includegraphics[width=.6\linewidth]{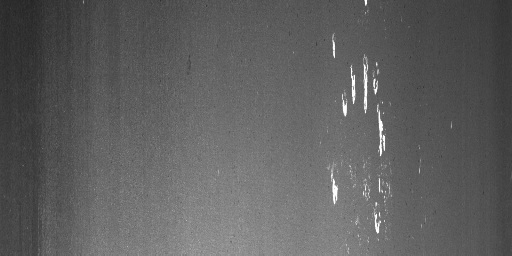}
  \caption{Severstal Steel Defect Dataset}
  \label{fig:sfig2}
\end{subfigure}
\begin{subfigure}{.5\textwidth}
  \centering
  \includegraphics[width=.6\linewidth]{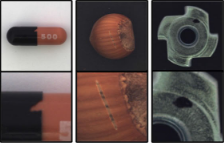}
  \caption{MVTec Anomaly Dataset}
  \label{fig:sfig2}
\end{subfigure}
\caption{Examples of items for each dataset}
\label{fig:fig}
\end{figure}


\subsection{The DAGM dataset}

The DAGM dataset \cite{Wieler2007} was published as part of the 2007 DAGM symposium (\textit{Deutsche Arbeitsgemeinschaft für Mustererkennung e.V.}, the German chapter of the IAPR (International Association for Pattern Recognition)). The DAGM and GNNS (German chapter of the European Neural Network Society) proposed a competition on the subject of \textit{Weakly Supervised Learning for Industrial Optical Inspection}, in order to help improve visual defect detection algorithms in industrial settings.

The dataset is divided in two parts. The first part, containing 6 classes, is commonly referred to as the \textit{development} dataset. The second part contains 4 classes and is referred to as the \textit{competition} dataset. 

Each development class contains 1000 non-defective images and 150 defective images. Each competition dataset contains 2000 non-defective images and 300 defective images. All images are 512$\times$512 pixel 8-bit grayscale. Defects are localized by an ellipse surrounding its precise location.

Each class is also divided in a training and a testing sub-dataset. They each contain half of the data, but the number of defective and non-defective items is not balanced between both sets.

The dataset was generated artificially, but was designed to resemble real defect detection problems in the manufacturing industry.
Each image contains at most one defect, whereas real products can have several defects. The images (both texture and defect) are artificially generated using statistical methods.

This dataset is among the most widely used datasets in industrial defect detection. However, it is an artificially generated dataset, and as such exhibits some statistical bias compared to real production data.

For example, Bozic et al. \cite{Bozic2021a} have proposed an algorithm that solves all 10 classes of the DAGM test set.


\subsection{NEU Surface Defect Database}

The Northeastern University Surface Defect Database was published in 2013 by Song et al. \cite{Song2013}. The dataset is constructed using greyscale pictures of defective hot-rolled steel sheets.
The 1800 images are 200$\times$200 pixels in size and are categorized into six defect types: crazing, inclusion, patches, pitted surfaces, rolled-in scales, and scratches. The defects are localized by a bounding box.
This dataset accurately classifies defects according to their type, which makes it interesting for evaluating defect type classifiers. However, it is also one of the smallest datasets, both in terms of image size and sample size. Furthermore, this dataset does not contain dedicated testing data, which makes it impossible to compare two metrics of algorithms tested on this dataset since the testing datasets will contain different items.

\subsection{Severstal Steel Defect Dataset}

The Severstal Steel Defect Dataset was published on Kaggle in 2018 \cite{Severstal2019} by PAO Severstal, a major Russian steel manufacturing company\footnote{\url{https://www.severstal.com/eng/}}.
This dataset contains 12,572 pictures of 512$\times$256 pixels representing  defective steel sheets and surfaces. The original Kaggle competition also featured a number of test images but their labels have never been published.
This dataset has the particularity of differentiating four types of defects, but the typology of these defects is not explained.
In addition, every defect is located pixel-wise on the image.

Very little detail was given about the creation of this dataset. Some images that do not have any defects show marks or peeling textures. Under these conditions, detecting defects can be a very difficult task because even non-defective items can vary a lot in appearance. We can also ask how the choice between defective and non-defective items was made, and why this choice makes sense in industry. This ambiguity has already been raised in the literature \cite{Bozic2021a}.


\subsection{KolektorSDD}

KolektorSDD is a dataset published in 2020 by Tabernik et al. \cite{Tabernik2020}. The dataset is constructed using images of defected electrical commutators. More specifically, the photographs depict defect of the plastic embedding of electrical commutators.

The dataset contains 399 images, of which 52 images contain a defect and 347 do not contain defects.
Defects are precisely located by a pixel-wise mask that covers the exact shape of the crack.
Images are grey-scale, 500 pixels wide and 1240 to 1270 pixels high. The authors recommend resizing the images to 512$\times$1408 pixels. \footnote{See \url{http://www.vicos.si/Downloads/KolektorSDD}}

This dataset was collected using real defective items, but no precise information was given on the collection method (type of camera, environment, lighting, etc.). These parameters could have a significant influence on the quality of the photographs and the results of learning algorithms. For example, Bozic et al. \cite{Bozic2021} have solved the test set of this dataset.


\subsection{KolektorSDD2}

KolektorSDD2 is a dataset published in 2021 by Bo\v{z}i\v{c} et al. \cite{Bozic2021a}. The dataset is constructed using color photographs of defective production items: as such, it is a different dataset than KolektorSDD.

The dataset contains 3335 images of approximately 230$\times$630 pixels. Of these images, 356 contain visible defects and 2979 are not defective. The dataset is already split in a training set containing 246 defective images and 2085 non-defective images, and a test set containing 110 defective images and 894 non-defective images.

Defects are localized using pixel-wise masks covering the exact shape of the defect.
The defects vary in shape, color, size and type (scratches, spots, imperfections, etc.)

Again, no information on the precise data collection conditions were given. However, this dataset is much more complex than its predecessor (more types of defects, addition of color channels, more types of objects). To this day, no method is able to successfully solve this dataset.


\subsection{MVTec Anomaly Dataset}

The MVTec Anomaly Dataset was published by MVTec Software in 2019 \cite{Bergmann2019,Bergmann2021}. This dataset contains 5354 images of different types of objects and textures.

The dataset presents multiple categories of objects: carpet, grid, leather, tile, wood, bottle, cable, capsule, hazelnut, metal nut, pill, screw, toothbrush, transistor, zipper.

The pictures are color images, with sizes between 700$\times$700 and 1024$\times$1024 pixels.

Each category is split into a testing set and a training set, like the DAGM dataset. The number of items in each group is different in each category. In total, the are 3629 training images and 1725 testing images. Moreover, all training images depict non-defective items.

This dataset is particularly suitable for training non-supervised algorithms, because of the absence of defective items in the training  dataset. It also takes a very different form than the previous datasets, because of the inclusion of non-textured objects such as screws, metal nuts, or even hazelnuts.


\subsection{Dataset comparison}

This section collates the characteristics of each dataset and discusses their main advantages and limitations for industrial defect detection.

\begin{itemize}
    \item The \textbf{DAGM Dataset} is one of the most popular datasets available in industrial defect detection. Since 2007, it has been used in dozens of articles \cite{Bozic2021a,Zavrtanik2021a,Siebel2008}. It contains 16,100 images in 10 classes which vary in terms of texture and defect type. The defects are annotated using bounding boxes, which can make segmentation harder since the bounding boxes contain non-defective areas. The dataset is artificially generated to resemble small metallic surfaces.
    \item The \textbf{NEU Surface Defect Database} is a popular defect dataset \cite{Dai2021,Cohn2021}. It contains 1,800 images, which makes this dataset rather small, but has the advantage of categorizing the different defect types. This dataset, showing real production line protographs, depicts manufactured steel sheets. However, this dataset does not contains defect-free items, and its lack of a default testing / training makes objectively comparing performances on this dataset impossible. 
    \item The \textbf{Severstal Steel Defect Dataset} is one of the largest industrial defect detection datasets, with 26,664 images. It has quickly gained popularity since its publication in 2019 \cite{Bozic2021a,Akhyar2021Detectors++:System}. It shows large images of steel sheets with large variations in appearance, and classifies its defects into four classes. However, the typology of the defects are not given, which makes the classification task seem arbitrary and difficult to interpret, and might even be the result of annotation errors \cite{Bozic2021a}. Furthermore, this dataset does not contain the ground truths of the testing data, which makes only the training data usable for both training and testing and makes objective performance comparison impossible. 
    \item The \textbf{KolektorSDD} dataset is a small dataset of 399 grayscale images showing plastic embeddings of electrical commutators. It was published in 2020 and has since gained some popularity \cite{Li2020,Lei2021}. This dataset has mostly been superseded by KolektorSDD2, which is very similar, although KolektorSDD2 does not contain the items of KolektorSDD.
    \item The \textbf{KolektorSDD2} dataset is a dataset of 3,335 color images showing defective production items. It was published in 2021 but has not yet reached widespread popularity, being used in less than a dozen articles to this day \cite{Bozic2021a,Lei2021}. This dataset, although larger than its previous version, is still rather small, and shows very few examples of defective items (356 in both the training and testing sets). Furthermore, the nature of the depicted items is unknown, unlike KolektorSDD.
    \item The \textbf{MVTec Anomaly Dataset} is a dataset of 5,354 images in 15 classes. This dataset was published in 2019 and its popularity is growing rapidly, both in the industrial defect detection community \cite{Zavrtanik2021a,Roth2021TowardsDetection} as well as in the wider computer vision community \cite{Venkataramanan2020} due to its great diversity of classes. These classes represent varied objects, from carpet textures to photographs of manufactured screws and bolts and even toothbrushes and hazelnuts. Not all classes of objects depict metallic or plastic manufactured items : however, this is the only dataset that proposes photographs of discrete manufactured items as opposed to textured items. These categories represent a small fraction of this dataset, which means that this dataset contains few images coming from industrial visual inspection. Furthermore, this dataset does not contain defective items in its training sets, meaning that only unsupervised training is possible.
\end{itemize}

Table \ref{table:datasets} compares the use-cases of the datasets. This comparison was made by analyzing the properties of the datasets.

\vspace{-0.7cm}

\begin{table}
    \centering
    \caption{Comparison of the described datasets:
    Detection is for datasets suitable for defect detection (Y if true, N otherwise);
    Classification is for datasets suitable for defect classification (Y if true, N otherwise);
    Labeling describes whether the defect boundaries are localized by a bounding box or a pixel-wise mask;
    Training defects describes the presence of defects in the training set (Y if present, N if not).
    }
    \label{table:datasets}
    \begin{tabular}{lccccccc}
        \toprule
        Dataset & Detection & Classification & Segmentation & Training defects \\ \midrule
        
        DAGM\textsuperscript{1} \cite{Wieler2007} & Y & N & Bounding box & Y \\ 
        
        NEU\textsuperscript{2} \cite{Song2013} & N & Y & Bounding box & Y \\ 
        
        Severstal Steel\textsuperscript{3} \cite{Severstal2019} & Y & Y & Pixel-wise mask & Y \\ 
        
        KolektorSDD\textsuperscript{4} \cite{Tabernik2020} & Y & N & Pixel-wise mask & Y \\ 
        
        KolektorSDD2\textsuperscript{5} \cite{Bozic2021a} & Y & N & Pixel-wise mask & Y \\ 
        
        MVTec\textsuperscript{6} \cite{Bergmann2019} & Y & Y & Pixel-wise mask & N \\ 
                 \bottomrule \\[-2.25ex]
        \multicolumn{6}{@{}p{0.9\textwidth}@{}}{\textsuperscript{1}\footnotesize{\url{https://hci.iwr.uni-heidelberg.de/content/weakly-supervised-learning-industrial-optical-inspection}}} \\
        \multicolumn{6}{@{}p{0.9\textwidth}@{}}{\textsuperscript{2}\footnotesize{\url{http://faculty.neu.edu.cn/songkc/en/zdylm/263265}}} \\
        \multicolumn{6}{@{}p{0.9\textwidth}@{}}{\textsuperscript{3}\footnotesize{\url{https://www.kaggle.com/c/severstal-steel-defect-detection}}} \\
        \multicolumn{6}{@{}p{0.9\textwidth}@{}}{\textsuperscript{4}\footnotesize{\url{https://www.vicos.si/resources/kolektorsdd/}}} \\
        \multicolumn{6}{@{}p{0.9\textwidth}@{}}{\textsuperscript{5}\footnotesize{\url{https://www.vicos.si/resources/kolektorsdd2/}}} \\
        \multicolumn{6}{@{}p{0.9\textwidth}@{}}{\textsuperscript{6}\footnotesize{\url{https://www.mvtec.com/company/research/datasets/mvtec-ad/}}} 
    \end{tabular}
\end{table}

\vspace{-1cm}

\section{Proposed guidelines and conclusion}
\label{section:discussion}

We reviewed the industrial needs for defect detection and defined a number of metrics that are used to evaluate an algorithm, both for segmentation and detection / classification. It is important that these metrics can be compared fairly between different articles when testing against the same benchmark. Therefore, we consider that it is important that a dataset be published with a default training / testing set split, so that tests can be run against the same items and are thus made comparable, and that algorithms be trained and tested using these datasets. Such datasets include DAGM or KolektorSDD.

Furthermore, it is important to consider the precision of the labeling. Labeling can indicate the typology of the defect by categorizing different defect types, and segment the defect using more or less precise indicators, namely bounding boxes and pixel-wise masks. Datasets which use more accurate labeling, and display a clear defect typology, such as MVTec or Severstal Steel, with defect classification and pixel-wise segmentation, should be adopted.

Real-world production datasets should also be preferred because they can exhibit some variance due to the production setting, such as ambient lighting, temperature, or camera settings. Testing algorithms against these real-world datasets give results that are likely to better represent the capabilities of the algorithm on a production line, and their robustness to disturbances. Only DAGM is generated artificially.

However, Table \ref{table:datasets} shows that no single benchmark, apart from MVTec which can only be trained using non-supervised algorithms, satisfies all desiderata. Therefore, the adequate benchmark must be chosen depending on the task at hand: namely, the classification or segmentation task that we wish to solve, and the supervision level of the algorithm.

We advocate the creation of a new benchmark which satisfies all criteria at once, namely the presence of defective and non-defective items in training and testing sets, the typology of defects for defect classification, and the segmentation of the defects using pixel-wise masks to allow training of supervised segmentation algorithms. The MVTec Dataset comes closest to this benchmark's requirements, but does not contain defective items in the training set.

\vspace{-0.3cm}
\section*{Acknowledgements}
\label{section:acknowledgements}

This work was carried out with the support of the Agence Nationale de la Recherche through the TEMIS ANR-20-CE10-0004 project.
\vspace{-0.3cm}
\bibliography{corrected_biblio}
\bibliographystyle{splncs04}

\end{document}